\documentclass[10pt,twocolumn,letterpaper]{article}

\usepackage{cvpr}              

\usepackage{url}
\usepackage{graphicx}
\usepackage{amsmath}
\usepackage{amssymb}
\usepackage{booktabs}
\usepackage{multirow}
\usepackage[dvipsnames]{xcolor}
\usepackage[rightcaption]{sidecap}
\usepackage{graphics}
\usepackage{colortbl}

\usepackage[accsupp]{axessibility}  

\usepackage[pagebackref,breaklinks,colorlinks, citecolor=blue]{hyperref}

\usepackage[capitalize]{cleveref}
\crefname{section}{Sec.}{Secs.}
\Crefname{section}{Section}{Sections}
\Crefname{table}{Table}{Tables}
\crefname{table}{Tab.}{Tabs.}

\def\cvprPaperID{10885} 
\def\confName{CVPR}
\def\confYear{2023}

\newcommand{\koushik}{\textcolor{black}}

\begin{document}

\title{\vspace{-0.5em}Evading Forensic Classifiers with Attribute-Conditioned Adversarial Faces\vspace{-0.5em}}

\author{Fahad Shamshad \quad
Koushik Srivatsan \quad
Karthik Nandakumar
\\
Mohamed bin Zayed University of AI, UAE \\
{\tt\small \{fahad.shamshad, koushik.srivatsan, karthik.nandakumar\}@mbzuai.ac.ae}
}

\maketitle

\begin{abstract}
The ability of generative models to produce highly realistic synthetic face images has raised security and ethical concerns. As a first line of defense against such fake faces, deep learning based forensic classifiers have been developed. While these forensic models can detect whether a face image is synthetic or real with high accuracy, they are also vulnerable to adversarial attacks. Although such attacks can be highly successful in evading detection by forensic classifiers, they introduce visible noise patterns that are detectable through careful human scrutiny. \koushik{Additionally, these attacks assume access to the target model(s) which may not always be true.} Attempts have been made to directly perturb the latent space of GANs to produce adversarial fake faces that can circumvent forensic classifiers. In this work, we go one step further and show that it is possible to successfully generate adversarial fake faces with a specified set of attributes (e.g., hair color, eye size, race, gender, etc.). To achieve this goal, we leverage the state-of-the-art generative model StyleGAN with disentangled representations, which enables a range of modifications without leaving the manifold of natural images. We propose a framework to search for adversarial latent codes within the feature space of StyleGAN, where the search can be guided either by a text prompt or a reference image. \koushik{We also propose a meta-learning based optimization strategy to achieve transferable performance on unknown target models.}
Extensive experiments demonstrate that the proposed approach can produce semantically manipulated adversarial fake faces, which are true to the specified attribute set and can successfully fool forensic face classifiers, while remaining undetectable by humans. Code: 
\url{https://github.com/koushiksrivats/face_attribute_attack}.
\end{abstract}

\vspace{-1em}
\section{Introduction}
\label{sec:intro}

Recent advances in deep learning such as generative adversarial networks (GAN) ~\cite{goodfellow2014generative,saxena2021generative} have enabled the generation of highly realistic and diverse human faces that do not exist in reality~\cite{thispersondoesnotexist}. While these synthetic/fake faces have many positive applications in video games, make-up industry, and computer-aided designs~\cite{gui2021review}, they can also be misused for malicious purposes causing serious security and ethical problems. For instance, recent news reports have revealed the potential use of GAN-generated face images in the US elections for creating fake social media profiles to rapidly disseminate misinformation among the targeted groups~\cite{ohlin2021defending,cai2021generative}. In a similar incident, a 17-year-old high school student successfully tricked Twitter into verifying a fake face profile picture of a US Congress candidate \cite{CNNBusinessNewsArticle} using a powerful generative model called StyleGAN2~\cite{karras2020analyzing}.



To combat this problem of synthetically-generated fake faces using GANs, several methods have been proposed to distinguish fake GAN-generated faces from real ones~\cite{juefei2022countering,wang2022gan}. The results reported in these works tend to indicate that \textit{simple, supervised deep learning-based classifiers are often highly effective at detecting GAN-generated images}~\cite{wang2020cnn,zhang2019detecting}. Such classifiers can be referred to as \emph{forensic classifiers/models}. However, a smart attacker may attempt to manipulate these fake images using tools from adversarial machine learning~\cite{akhtar2021advances} to bypass forensic classifiers, while maintaining high visual quality. A recent attempt in this direction is the work of Li \textit{et al.}~\cite{li2021exploring}, which demonstrates that 
adversarially exploring the manifold of the generative model by optimizing over the latent space can generate realistic faces that are misclassified by the target forensic detectors. Furthermore, they show that the resulting adversarial fake faces contain fewer artifacts compared to traditional norm-constrained adversarial attacks in the image space.

\begin{figure*}[t]
\centering
\includegraphics[width=0.95\textwidth]{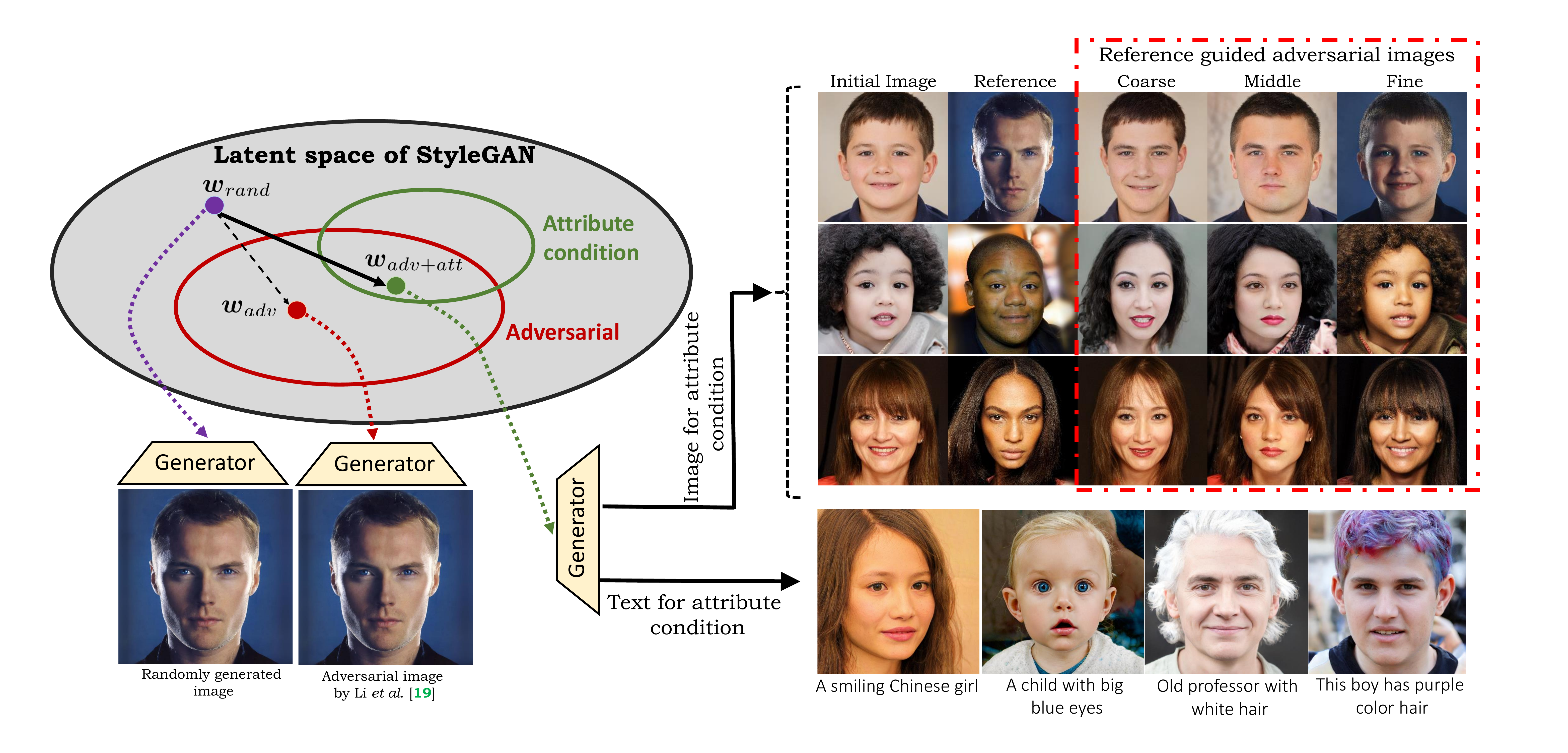}
\caption{\textit{Left:} An illustration of our attribute-conditioned adversarial face image generation approach. Instead of generating face image randomly ({\color{Plum}purple dot}) and making it adversarial using approach of Li \textit{et al.}~\cite{li2021exploring} ({\color{red}red dot}), we leverage the highly disentangled latent space of the StyleGAN2 to generate attribute-conditioned adversarial face images ({\color{ForestGreen}green dot}) that are misclassified by the target forensic classifier. \textit{Right:} Generated adversarial face images via our proposed image-driven and text-guided method.}
\label{fig:ip_pipeline}
\end{figure*}

While being the first to introduce the generative model based adversarial attack to fool \textit{forensic face classifiers}, the work of Li \textit{et al.}~\cite{li2021exploring} suffers from one key limitation: the inability to control the attributes of the generated adversarial faces like skin color, expression, or age. This kind of control over face attributes is essential for attackers to rapidly disseminate false propaganda via social media to specific ethnic or age groups. Therefore, it is imperative for image forensics researchers to study and develop these attribute-conditioned attacks to expose the vulnerabilities of existing forensic face classifiers, with the eventual goal of designing effective defense mechanisms in the future.   

In this paper, we propose a method to generate realistic image-driven or text-guided adversarial fake faces that are misclassified by forensic face detectors. Our method leverages the highly disentangled latent space of StyleGAN2 to craft attribute-conditioned unrestricted attacks in a unified framework, as illustrated in Figure \ref{fig:ip_pipeline}. 
Specifically, for generating a fake face that matches the attributes of a given reference image, we propose an efficient algorithm to optimize attribute-specific latent variables in an adversarial manner. This results in an effective transfer of the desired coarse or fine-grained attributes from the reference image to the generated fake image. For generating fake faces that match given text descriptions, we leverage the joint image-text representation capabilities of Contrastive Language-Image Pre-training (CLIP)~\cite{radford2021learning} to enforce consistency between the generated adversarial face image and text description. Our contributions can be summarized as follows:

\begin{itemize}
\item We propose a framework to generate adversarial fake faces with a specific set of attributes defined using a reference image or a text prompt. This is achieved by performing both semantic manipulations and adversarial perturbations to a randomly generated initial image in the latent space of StyleGAN2~\cite{karras2020analyzing}. For \textbf{image-based attribute conditioning}, we transfer semantic attributes from the given reference image by searching the adversarial space guided by perceptual loss. For \textbf{text-based attribute conditioning}, we take advantage of multimodal CLIP~\cite{radford2021learning} and use its text-guided feature space to search for adversarial latent codes within the feature space of StyleGAN2 to fool forensic classifiers. 

\item {\koushik{We leverage a meta-learning-based optimization strategy to generate adversarial images that are more transferable to the unknown black-box forensic classifiers compared to ensemble-based approach. }}
\item We show that semantic changes (\eg hair color changes, manipulating eye sizes, \etc) introduced by our method appears benign to human subjects while being adversarial to deep forensic classifiers. We further validate this claim by measuring the perceptual similarity between the original and adversarial samples. We also conduct experiments to show that faces generated using the proposed approach not only exhibit the desired attributes but are also able to fool the target forensic classifier with high success rate.
\end{itemize}
At the outset, we would like to strongly emphasize that our goal is \emph{not to circumvent a face recognition system}. Our only objective is to generate realistic synthetic faces with specific attributes that can evade a forensic classifier, which is trained to detect synthetic images. The reference image or text prompt merely guides the generation process towards the required attributes. Hence, \emph{we make no explicit attempt to preserve the identity of either the randomly generated initial image or the reference image in the final generated image}. \koushik{However, we experimentally show that, it is also possible to preserve the identity of the generated image without sacrificing its adversarial nature. }



\section{Related Work}
\label{sec:related_work}

\textbf{StyleGAN:} StyleGAN~\cite{karras2019style} has attracted a lot of attention in the vision and graphics community due to its highly disentangled latent spaces and impressive image generation quality.
StyleGAN is a generator architecture based on generative adversarial networks, which borrows interesting properties from the style transfer literature. While various variations of this architecture exist, we exclusively focus on StyleGAN2~\cite{karras2020analyzing}. Formally, the StyleGAN architecture is composed of two networks. A mapping network $\mathcal{M}$ takes a random latent vector $\textbf{z} \sim \mathcal{N} (0, 1)$ (which is not necessarily disentangled) and projects it into a disentangled latent vector $\boldsymbol{\omega} = \mathcal{M}(\textbf{z}) \in \mathcal{W}$. A synthesis network $\mathcal{G}$ synthesizes an image $\boldsymbol{I}$ from the disentangled latent vector $\boldsymbol{\omega}$, i.e., $\boldsymbol{I} = \mathcal{G}(\boldsymbol{\omega})$. The disentangled latent vector $\boldsymbol{\omega}$ impacts image generation at different spatial resolutions, and this allows us to control the synthesis of an image. In particular, we can apply the style of one image to another by mixing the disentangled latent vectors of these images together. In the context of face generation, it has been shown that the initial layers of $\mathcal{G}$ (corresponding to coarse spatial resolutions) control high-level aspects such as pose, general hairstyle, and face shape, while the middle layers control smaller-scale facial features such as hairstyle, and final layers (fine spatial resolutions) primarily control the color scheme~\cite{richardson2021encoding}.

\textbf{Latent Space Manipulation:} Generative models have been widely used for the editing of images in the latent space~\cite{zhu2020domain}. Due to its highly disentangled latent space, considerable effort has been made to leverage StyleGAN2 for such tasks~\cite{bermano2022state}. A plethora of approaches have been proposed to find meaningful semantic directions in the latent space of the StyleGAN2.
These range from full-supervision in the form of facial priors~\cite{tewari2020pie} or semantic labels~\cite{abdal2021styleflow} and to unsupervised approaches~\cite{voynov2020unsupervised}. Few works have also explored self-supervised approaches that include local image editing via mixing of latent codes~\cite{kafri2021stylefusion}, and the use of contrastive language-image (CLIP) for text-prompt guided image editing~\cite{patashnik2021styleclip}. Different from these works, in this paper, we aim to find the \textit{adversarial} face images that lie on the manifold of the StyleGAN with the desired attributes.

\textbf{Image Forensics and Adversarial Attacks:} Researchers have proposed a variety of deep learning-based methods to detect images generated through GANs~\cite{guo2022eyes,wang2020cnn,yu2018attributing}. This line of work has found that simple supervised forensic classifiers effectively detect GANs generated images. These forensic models generally act as the first line of defense against \textit{Deepfakes}. However, these models can be easily fooled by carefully crafting adversarial examples using tools from adversarial machine learning~\cite{akhtar2021advances}. Several defense mechanisms against adversarial attacks have been proposed that include adversarial training~\cite{madry2017towards}, distillation~\cite{papernot2016distillation}, and self-supervised approaches~\cite{naseer2020self,chen2020adversarial}.

\textbf{Unrestricted Adversarial Examples.} The images that successfully fool classifiers without confusing humans are referred as unrestricted adversarial examples (UAEs)~\cite{song2018constructing}. To generate UAEs, one can apply various modifications to the original image such as attribute editing~\cite{qiu2020semanticadv}, spatial transformation~\cite{xiao2018spatially}, etc. Although recent works \cite{nguyen2022improving, xiang2022egm, chen2021unrestricted} show promising results, none of them allow fine-grained attribute control while generating UAEs. Attempts have been made to create UAEs using pre-trained generative models via optimizing over their latent space~\cite{song2018constructing, li2021exploring}. These works show that the generated adversarial examples contain less artifacts as compared to traditional norm-constrained adversarial attacks in the image space. However, they do not provide the ability to control the attributes of the generated adversarial faces like skin color, expression, or age.
Recently, \cite{jia2022adv} focuses on fooling the face recognition models by generating adversarial images with specific attributes.
However, they are restricted to only 5 fixed attributes unlike our framework, which offers improved flexibility to manipulate multiple attributes using image and text-driven approaches.

\section{Proposed Method}

Our goal is to generate semantically meaningful attribute-conditioned face images that can fool the forensic classifier. To achieve this objective, we directly manipulate the latent space of StyleGAN to incorporate specific face attributes in a controlled manner. Let $\mathcal{G}_L$ denote the synthesis network of the trained StyleGAN with $L$ layers. The input $\boldsymbol{\omega}$  to $\mathcal{G}_L$ in the extended latent space $\mathcal{W}^{+}$ is formed by stacking $L$ copies of the output $\boldsymbol{\omega_m} \in \mathbb{R}^{1 \times 512}$ of the mapping network $\mathcal{M}$ i.e., $\boldsymbol{\omega} = [\boldsymbol{\omega_m},\boldsymbol{\omega_m},...,\boldsymbol{\omega_m}]^{T} \in \mathbb{R}^{L \times 512}$. More specifically, the $l^{th}$ latent code (row) of $\boldsymbol{\omega}$ represents the input to the $l^{th}$ layer of $\mathcal{G}_L$, and controls the $l^{th}$ level of detail of the generated image, $l = 1,2,\cdots,L$. Changing the value of a certain layer code will change the attribute associated with that layer. In addition to $\boldsymbol{\omega}$, $\mathcal{G}_L$ also takes a collection of latent noise vectors $\boldsymbol{\eta}$ as input. Here, $\boldsymbol{\eta}$ represents uncorrelated Gaussian noise that controls the stochastic variations of the generated face at each layer. We seek to generate a synthetic image $\textbf{I} = \mathcal{G}_L(\boldsymbol{\omega},\boldsymbol{\eta})$ by crafting $\boldsymbol{\omega}$ and $\boldsymbol{\eta}$ in a semantically meaningful way so that it can fool the forensic classifier while capturing the desired attributes. 

The set of desired attributes to be included in the synthetically generated face can be specified in two ways. Firstly, a reference image may be available and the attacker wants to transfer some attributes (e.g., pose, expression, or color) of the reference face image to the generated image. Secondly, some reference text describing the desired attributes (e.g., \texttt{A smiling Chinese girl} or \texttt{A baby with big blue eyes}) may be available. Depending on how the desired attributes are specified, we propose two different methods to generate fake faces. In both these methods, we start with a random latent vector $\boldsymbol{z}$ and map it to an initial disentangled latent code $\boldsymbol{\omega}_s$. The difference between the two methods lies in how this initial code $\boldsymbol{\omega}_s$ is manipulated to obtain the final latent code $\boldsymbol{\omega}^{*}$ along with the noise vector $\boldsymbol{\eta}^{*}$. Since our objective is to generate a face image with specified attributes that can fool the forensic classifier, the adversarial capability of the generated image is also taken into account during the manipulation of the latent space and noise vector. This is in contrast to norm-based adversarial attacks, where a synthetic image is first generated and adversarial perturbations are added later in a separate step.

\subsection{Image as a Reference} \label{sec:image_ref}
\begin{figure}[t]
\centering
\includegraphics[width=0.48\textwidth]{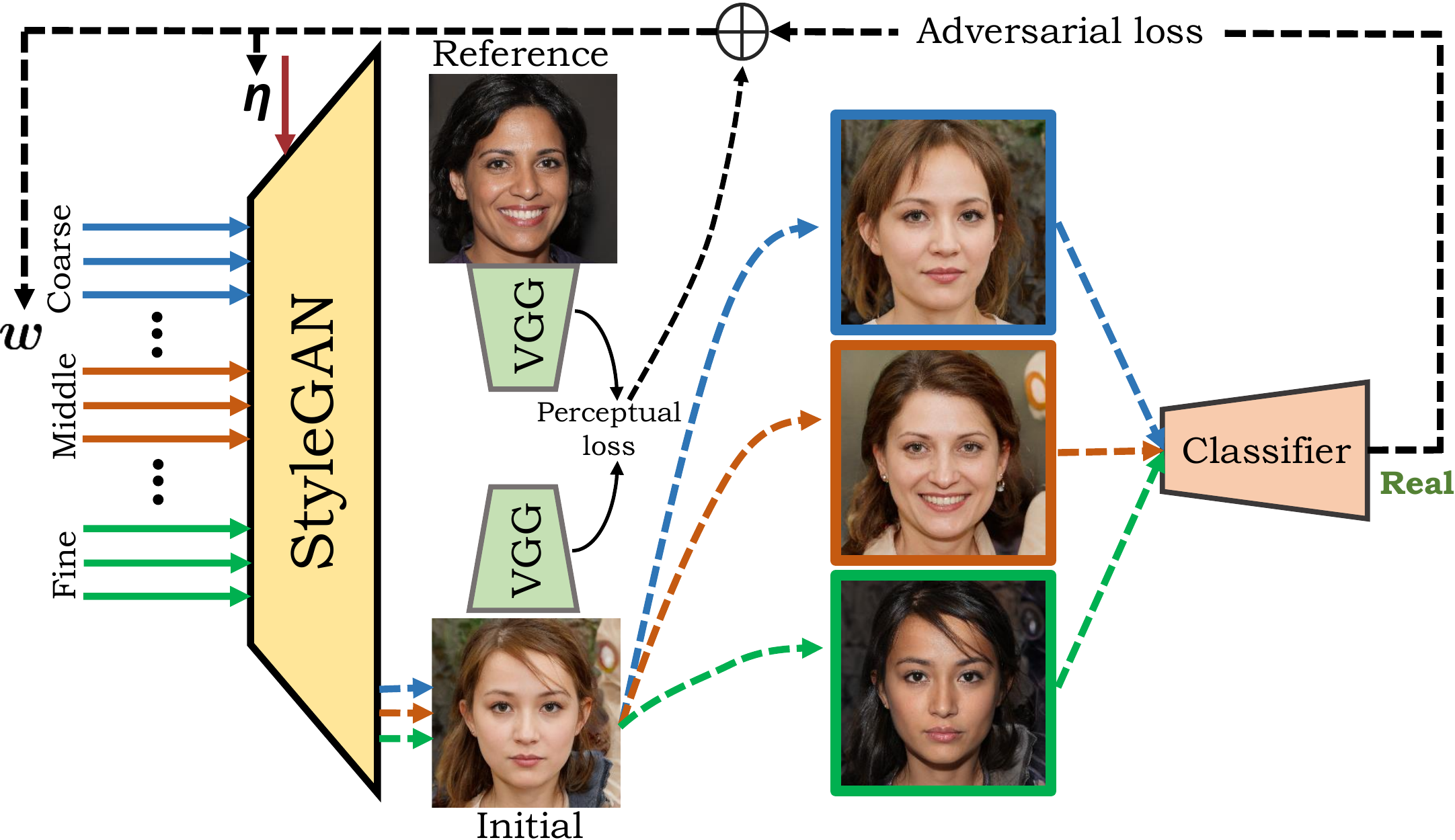}
\caption{The framework of the proposed attribute-conditioned image-driven approach. Given a reference image, we randomly generate an initial image from the StyleGAN generator. Depending on which attributes to transfer, we adversarially optimize latent codes over the attribute-specific layers ({\color{blue}coarse}, {\color{orange}middle}, {\color{green}fine}) of the generator to minimize the perceptual distance between the reference image and the initial image in the feature space of the pre-trained VGG-19 network.}
\label{fig:ref_img}
\end{figure}

In this scenario, a reference image $\textbf{I}_r$ is provided and the goal is to transfer the desired attributes from the reference image (pose, expression, or color) to the generated image. A naive way to achieve this goal would be to reverse engineer the latent code $\boldsymbol{\omega}_r$ of the reference image by projecting 
$\textbf{I}_r$ into the latent space of StyleGAN2, and replace the desired attribute-specific rows of the intermediate code $\boldsymbol{\omega}_s$ with the corresponding rows from $\boldsymbol{\omega}_r$. Though this approach can be effective, it requires an additional step of inverting the reference image into the StyleGAN2 latent space. While an optimization-based approach would require several minutes to achieve this inversion,  encoder-based inversion would require training of a separate encoder network to infer the latent codes of the reference image~\cite{xia2021gan}. Instead, we propose a more efficient algorithm to adversarially optimize only over the latent codes of the desired attribute-specific layers of the StyleGAN as shown in Figure \ref{fig:ref_img}. The proposed approach is order of magnitude faster than the naive optimization-based inversion approach and does not require any separate training of encoder network.

\begin{figure}[t]
\centering
\includegraphics[width=0.48\textwidth]{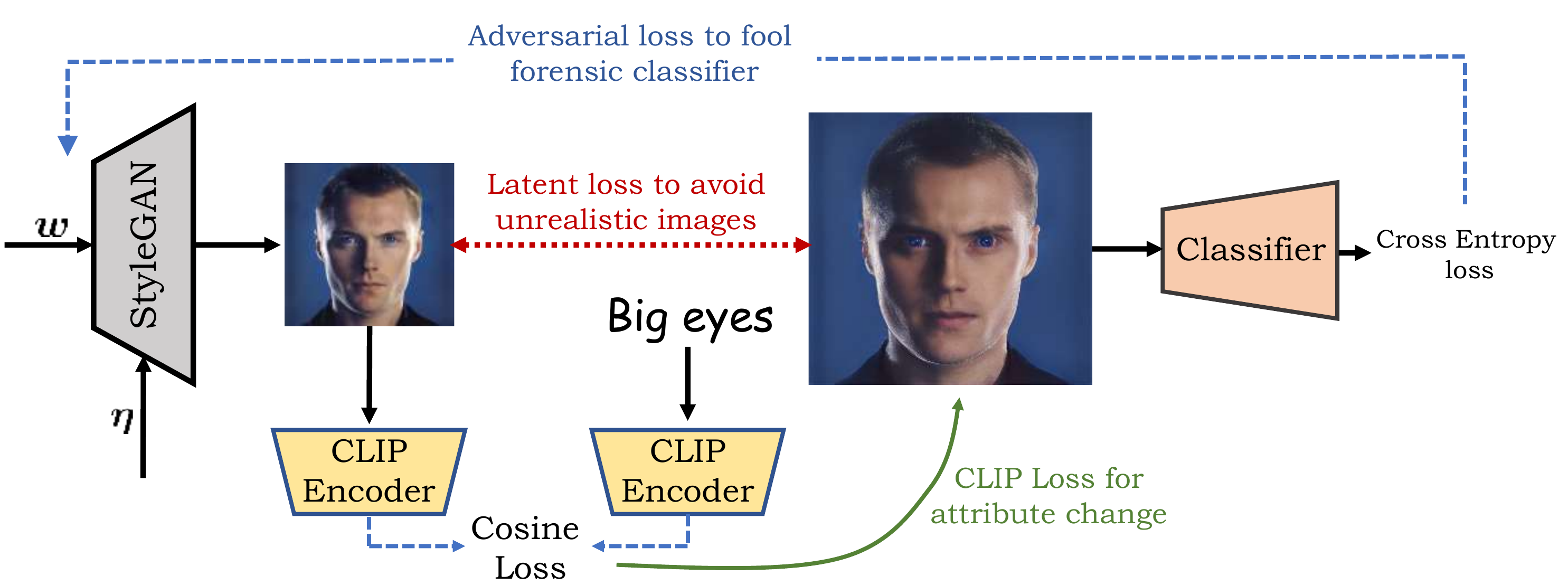}
\caption{The framework of the proposed attribute-conditioned text-driven approach. We adversarially optimize over latent codes and noise tensors of the StyleGAN generator to minimize the distance between the initial image embedding and reference text embedding in the CLIP space. Latent loss between the adversarial image and the initial image is enforced to avoid unrealistic images (see Section \ref{sec:text_ref} for details).}
\label{fig:ip_pipeline_clip}
\end{figure}

Specifically, we aim to minimize the following objective function:
\begin{align} \label{eq:1}
    (\boldsymbol{\omega}^*,\boldsymbol{\eta}^*) = \underset{\boldsymbol{\omega},\boldsymbol{\eta}}{\arg\min} \; \nonumber 
    \Vert \phi(\mathcal{G}_L(\boldsymbol{\omega},\boldsymbol{\eta})) - \phi(\textbf{I}_r) \Vert^{2}_2 \\
    + \lambda_1\Vert \boldsymbol{\omega}- \boldsymbol{\omega}_s \Vert^{2}_2 + \lambda_2\text{BCE}(\mathcal{C}(\mathcal{G}_L(\boldsymbol{\omega},\boldsymbol{\eta})),y=1), 
\end{align}
\noindent where the first term enforces the perceptual similarity between the reference image and the generated image, the second term ensures that the manipulated latent codes do not deviate much from the generated intermediate latent codes, and the third term is the adversarial loss (which ensures that the generated image is able to fool the forensic classifier). In Eq. (\ref{eq:1}), $\phi$ represents the layers of a pre-trained VGG-19 network and the perceptual loss is computed in the feature space of the VGG-19 network. Moreover, the latent codes $\boldsymbol{\omega}, \boldsymbol{\omega}_s \in \mathcal{W}^+$ and the second term in Eq. (\ref{eq:1}) ensures that the adversarial latent codes lie close to the StyleGAN manifold, thereby avoiding the possibility of generating unrealistic face images. Finally, $\lambda_1$ and $\lambda_2$ are hyperparameters taking positive values, BCE represents binary cross-entropy loss, $\mathcal{C}$ represents the forensic classifier that maps an image to an output $y \in \{0,1\}$, where $y=1$ represents the real class and $y=0$ represents the synthetic class. The objective function in Eq. (\ref{eq:1}) can be solved directly using gradient descent methods and the final adversarial fake face can be obtained as $\textbf{I} = \mathcal{G}_L(\boldsymbol{\omega}^{*},\boldsymbol{\eta}^{*})$.

\subsection{Text as a Reference} \label{sec:text_ref}

Our text-based attribute-conditioned adversarial face generator (shown in Figure \ref{fig:ip_pipeline_clip}) leverages the power of rich, joint vision-language representation learned by the CLIP model. Specifically, our optimization scheme aims to modify the latent vector of the StyleGAN in $\mathcal{W}^+$ space under CLIP loss to generate adversarial face images with attributes described by the text prompt. Given a text description $t$, we minimize the following objective function:
\begin{align} \label{eq:2}
    (\boldsymbol{\omega}^*,\boldsymbol{\eta}^*) & = \underset{\boldsymbol{\omega},\boldsymbol{\eta}}{\arg\min} \; \nonumber \mathcal{L}_\text{clip}(\mathcal{G}_L(\boldsymbol{\omega},\boldsymbol{\eta}),t) 
    + \lambda_1\Vert \boldsymbol{\omega}- \boldsymbol{\omega}_s \Vert^2_2 \\
    &+ \lambda_2\text{BCE}(\mathcal{C}(\mathcal{G}_L(\boldsymbol{\omega},\boldsymbol{\eta})),y=1), 
\end{align}
\noindent where $\mathcal{L}_\text{clip}(^.,^.)$ denotes the cosine distance between the embeddings of the generated image via StyleGAN $\mathcal{G}_L(\boldsymbol{\omega},\boldsymbol{\eta})$ and reference text prompt $t$. Note that eqs (\ref{eq:1}) and (\ref{eq:2}) are similar, except for the fact that the perceptual loss (first term in Eq.(\ref{eq:1})) is replaced by the CLIP loss (first term in Eq.(\ref{eq:2})) \koushik{Furthermore, to improve the efficiency of the search, $\boldsymbol{\omega}_s$ is used as the initialization for $\boldsymbol{\omega}$ rather than using a randomly initialized $\boldsymbol{\omega}$.}

Since CLIP has been trained on 400 million text-image pairs and text can express a much wider range of visual context as compared to images, the text-guided approach is generally able to produce adversarial faces with desired attributes that are not possible with image as reference images. It should be noted that the text prompts are not limited to a single attribute at a time, as will be shown in Section~\ref{sec:textasref}, where different combinations of hair attributes, straight/curly and short/long, are specified and each yields the expected outcome. This degree of control is difficult to achieve via the reference image-based approach. Furthermore, it can be argued that finding a reference image with the desired attributes is considerably more difficult for the attacker compared to providing a text prompt. Therefore, the text-guided approach is likely to be more useful in practice compared to the reference image guidance.

\subsection{\koushik{Meta Learning based Adversarial Attack}} \label{sec:metalearning}
\koushik{Recent methods to evade forensic classifiers \cite{li2021exploring}  generally assume access to the target model. The adversarial images generated by this approach do not generalize well to gradient-unknown (black-box) models, thus resulting in poor transferability. Although ensemble based methods have shown to generalize better, they still tend to overfit to the white box models. Inspired from the domain generalization literature \cite{zhou2022domain} and recent works in adversarial attacks \cite{yuan2021meta, yin2021adv}, we propose to optimize for the latents using meta-learning, by simulating both the white box and black box environments. Specifically, given a total of $T+1$ forensic classifiers, we randomly sample $T$ classifiers from them and use $T-1$ for meta-train and the remaining model for meta-test. For every iteration, we shuffle and choose different combinations of meta train-test pairs from the set of $T$ classifiers. The latents are first updated to evaluate on the meta-test model, and finally the aggregated losses from the meta-train and meta-test stages are used to optimize the latent for the current iteration. Detailed description of our meta objective function is given in the supplementary section. }

\begin{table*}[]
\caption{\textit{The attack success rate (ASR)} of the adversarial images generated by our image-driven and text-guided approaches along with the norm-constrained noise-based methods. The first row reports the clean accuracy of the forensic classifiers. In the subsequent rows, we report \textit{ASR} by optimizing over only the ``latent'' code and over both the latent code and the noise tensor (``noise and latent''). FID scores (lower is better) are averaged across all forensic classifiers.}
\centering
\resizebox{0.85\textwidth}{!}{
    \begin{tabular}{|c|c|c|c|c|c|c|} 
    \hline
    \multirow{2}{*}{\textbf{Method}} & \multicolumn{5}{c|}{\textbf{Models}} &  \multirow{2}{*}{\textbf{FID}~\cite{heusel2017gans}} \\ 
    \cline{2-6}
    \multicolumn{1}{|c|}{}                                  & ResNet-18 & ResNet-50 & VGG-19 & DenseNet-121 & Wang \textit{et al.}~\cite{wang2020cnn} & \\ 
    \hline
    Clean accuracy                                          & 94$\%$ & 97$\%$ & 96$\%$ & 96$\%$ & 81$\%$ & - \\ 
    PGD $L_{inf}$~\cite{madry2017towards}                   & {98$\%$ } & {100$\%$ } & {100$\%$ } & {95$\%$ } & {86$\%$ } & 49.54 \\ 
    FGSM $L_{inf}$~\cite{goodfellow2014explaining}          & {100$\%$ } & {100$\%$ } & {100$\%$ } & {100$\%$ } & {95$\%$ } &  38.24 \\
    Latent  (image)                                               &   {100$\%$ }       &    {100$\%$ }        &     {100$\%$ }    &   {100$\%$ }           &     {89$\%$ }    &  28.31  \\ 
    Noise and latent (image)                                     &    {100$\%$ }       &      {100$\%$ }       &     {100$\%$ }    &      {100$\%$ }        &      {100$\%$ }    &  26.44  \\              
    Latent (text)                                                &    {100$\%$ }       &    {100$\%$ }        &    {100$\%$ }    &       {100$\%$ }         &       {91$\%$ }    &  34.73 \\ 
    Noise and latent (text)                                      &    {100$\%$ }        &    {100$\%$ }        &   {100$\%$ }      &      {100$\%$ }         &        {100$\%$ }   &  31.92  \\
    \hline    
    \end{tabular}
    }
\label{tab:classifiers}
\end{table*}

\section{Experiments}
For all experiments, we use a generator of StyleGAN2 pretrained on the FFHQ face dataset~\cite{karras2019style} to generate adversarial face images. Following Li \textit{et al.}~\cite{li2021exploring}, to evaluate the effectiveness of the proposed approach, we train \koushik{ EfficientNet-B3 and Xception along with} VGG-19, ResNet-18, ResNet-50, and DenseNet-121 \koushik{networks,} as the forensic models to distinguish real faces from the generated fake faces. To train these forensic models, we use 50,000 real face images from the FFHQ dataset and 50,000 StyleGAN2 generated images. For training, we apply standard ImageNet augmentation: resize the face images to \koushik{$224 \times 224$}, and apply ImageNet normalization. We train classifiers with Adam optimizer using Binary Cross-Entropy loss with a learning rate of $2 \times 10^{-4}$ and batch size of 64. \koushik{All the classifiers achieve above $90\%$ accuracy on the hold-out test set (last $1,000$ images of FFHQ and last $1,000$ images from StyleGAN generated images) as shown in Table \ref{tab:classifiers}.}
We also attack the forensic classifier proposed by Wang \textit{et al.}~\cite{wang2020cnn}, which is capable of detecting images generated by nearly a dozen generative models, including StyleGAN2, with high accuracy. Specifically, they use ResNet-50 architecture to classify an image as real or fake via training it on $720,000$ images, half of which are real and half of which are generated using Progressive GAN~\cite{karras2017progressive}.

Note that although both GAN discriminator and forensic classifier have the same functionality (distinguish real and generated images), there are two key differences: (i) While the discriminator is learned alternately along with the generator during GAN training (discriminator is expected to be maximally confused at the end of GAN training), the forensic classifier is learned from scratch post-GAN training without any knowledge of the discriminator. (ii) Discriminator is controlled by an adversary aiming to generate synthetic images and is tied to a specific generator, but the forensic classifier is trained by a defender aiming to detect synthetic images. Forensic classifier can be trained using samples generated from multiple unrelated generators. For example, the forensic classifier by Wang \textit{et al.}~\cite{wang2020cnn} is trained on the images (including non-face images) generated via Progressive GAN.

Our goal is to break the aforementioned forensic classifiers using the proposed image-driven and text-guided semantic attacks. Since the final images are synthetic, there is no ground truth to compute SSIM and PSNR values. \koushik{The metrics to evaluate the performance are \textit{attack success rate (ASR)} (higher values indicate better attacks) and human evaluation. The experimental results for the white-box and black box setting are shown in Table \ref{tab:classifiers} and Table \ref{tab:blackbox_results} respectively. }
\begin{figure}[t]
\centering
\includegraphics[width=0.44\textwidth]{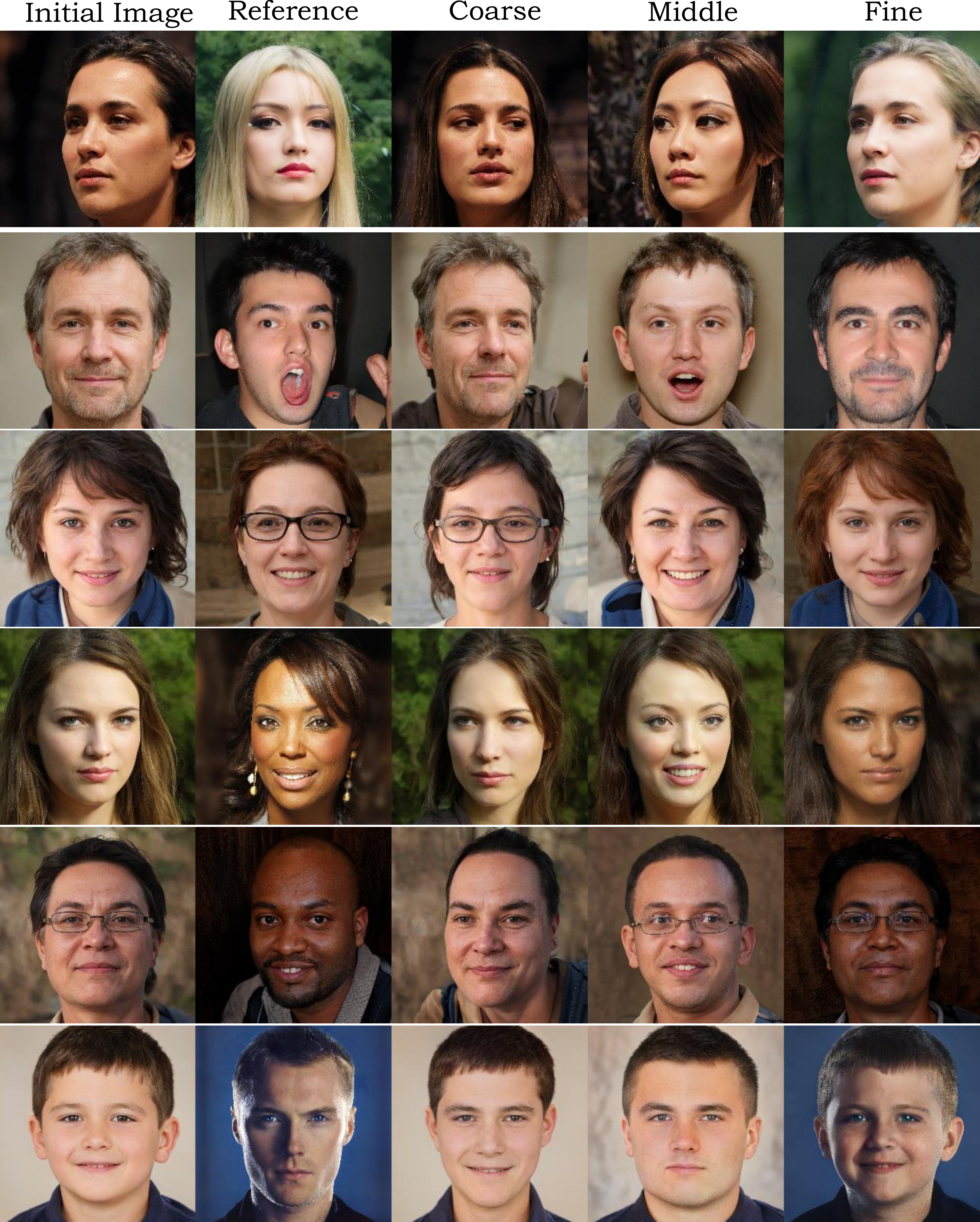}
\caption{Attribute-conditioned adversarial face images generated via proposed reference image based approach. The proposed approach faithfully transfer the desired attributes (coarse (pose), middle (expression), and fine (color)) from the reference image to the initial image by adversarially optimizing over specific layers of the pre-trained StyleGAN generator. All the generated images are misclassified by the forensic classifier (DenseNet-121). Additional results are provided in the supplementary material.}
\label{fig:qual_refimg}
\end{figure}
\begin{figure}[t]
\centering
\includegraphics[width=0.2\textwidth]{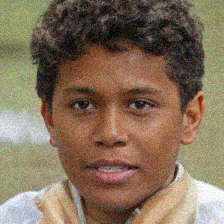}
\includegraphics[width=0.2\textwidth]{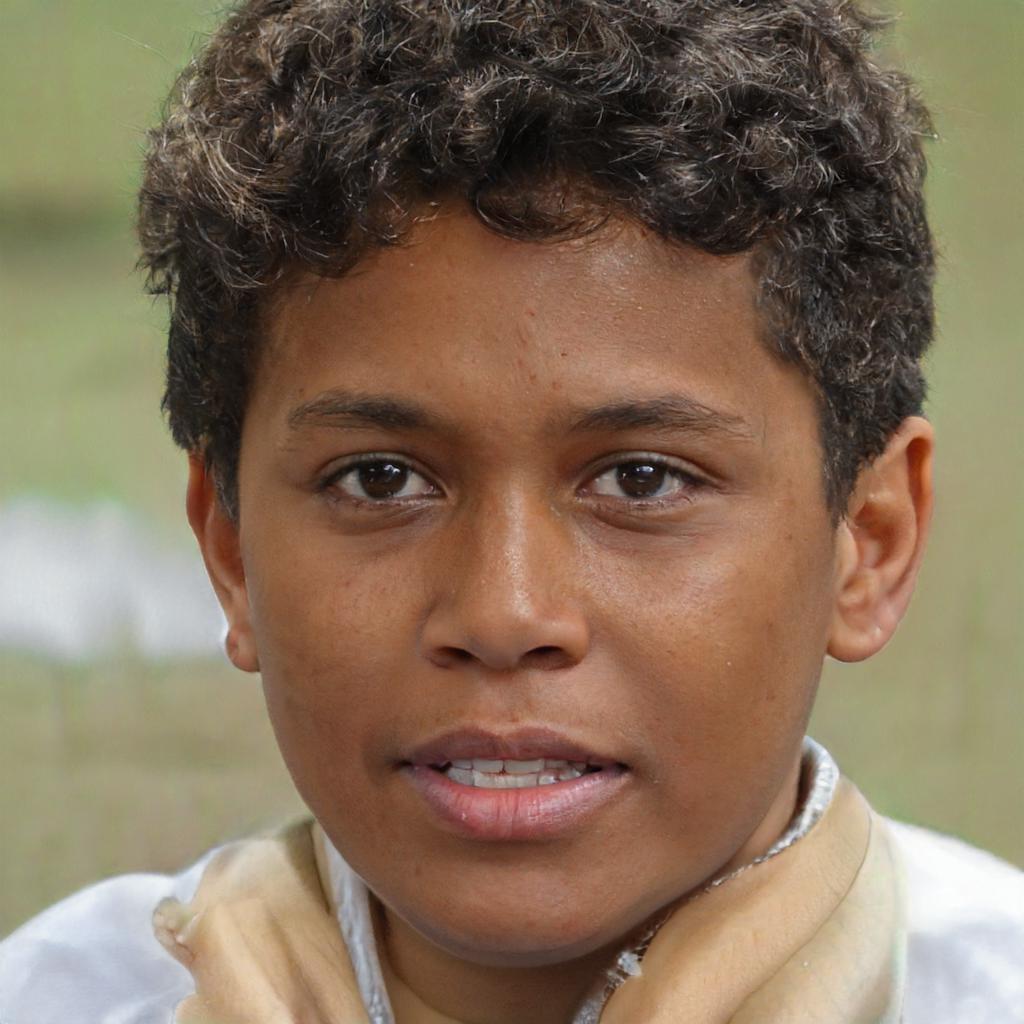}
\caption{ On left is the image generated by PGD~\cite{madry2017towards} attack respectively with perturbation 0.06. On the right is the image generated via the proposed image-based approach. Although these images can bypass the target forensic model (here DenseNet-121), but the image generated by the proposed approach (modified to two-stage for the comparison purpose where the attribute is transferred in the first stage and image is made adversarial via optimizing over latent in the second stage) contains fewer artifacts.}
\label{fig:qual_textguidedsupp_}
\end{figure}
\subsection{Image as Reference:} To evaluate our reference image-based approach, we randomly generate 100 images from the StyleGAN2 generator as our initial images and select 100 images from the CelebA-HQ dataset~\cite{karras2017progressive} as reference images. Note that the generator consists of $18$ style layers. We optimize the top 4 layers for coarse-level changes (such as pose), 5$^{th}$ to 8$^{th}$ layers for middle-level changes (such as expression), and the last $10$ layers from 9$^{th}$ to 18$^{th}$  for fine-level changes (such as color). We use Adam optimizer to jointly optimize noise and attribute-specific style vectors to minimize Eq. \ref{eq:1} with a learning rate of 0.01. We perform multiple gradient descent steps, usually between 20 to 30, until the target forensic classifier is fooled. We use features of the 16$^{th}$ layer of the pre-trained VGG-19 network to enforce perceptual similarity between the generated output and the reference image. For all experiments, we set value of $\lambda_2$ (the weight of BCE loss in Eq. \ref{eq:1}) to $0.005$. The value of $\lambda_1$ is high for top layers, as it requires a relatively large change in the pose of the initial image, and low for the later layers, which only account for the color transfer from the reference to the initial image. More details about the hyperparameter $\lambda_2$ are provided in the supplementary material. For noise-based methods, we chose FGSM~\cite{goodfellow2014explaining} and PGD~\cite{madry2017towards} attacks under $L_{inf}$ constraint on the target forensic models. The step size for these attacks is 0.01, with number of iterations set to 50. The maximum perturbation size is 0.06 (for normalized image pixel values between 0 and 1).

Visual attribute-conditioned adversarial examples are shown in Figure \ref{fig:qual_refimg} and the accuracies of the forensic classifiers for adversarial images are indicated in Table \ref{tab:classifiers}. Clearly, the proposed approach can faithfully transfer low-level, mid-level, and high-level attributes of the reference image to the generated image depending on the layers of the generator that we optimize. 
Moreover, the lower FID scores indicate that the proposed method generates more realistic synthetic faces with high visual quality (compared to PGD \cite{madry2017towards} and FGSM \cite{goodfellow2014explaining}), which can also be observed in Figure \ref{fig:qual_textguidedsupp_} where the adversarial image generated by the proposed approach contains fewer noise artifacts.  
\begin{SCtable}[\sidecaptionrelwidth][t]
\centering
\footnotesize
\caption{ Naive vs. proposed image-driven approach, averaged over 100 samples.} 
 \resizebox{0.25\textwidth}{!}{%
    \begin{tabular}{|c|c|c|}
    \hline
    & \textbf{Time (sec)} & \textbf{\textit{ASR}}\\
    \hline
    \textbf{Naive} & 105 & 100$\%$\\
    \hline
    \textbf{Proposed} & 23 & 100$\%$\\
    \hline
    \end{tabular}
 }
\label{tab:naive}
\end{SCtable}

\textbf{Comparison with Naive Approach:} As mentioned in Section \ref{sec:image_ref}, a naive approach to craft a reference image-based attack is to first invert the reference image in the latent space of StyleGAN via the GAN inversion method, leverage the style-mixing property to transfer desired attributes to the randomly generated initial image, and adversarially optimize over the latent code and noise. In Table \ref{tab:naive}, it can be seen that our proposed approach is about $5$ times faster than the naive approach and has a similar performance in terms of the adversarial attack success rate. Furthermore, compared to conventional encoder-based methods~\cite{tov2021designing}, our approach does not require training of extra encoder to infer the latent of reference image.
\begin{table}[]
\caption{Evaluation of the realism of the generated adversarial images via proposed and noise-based approaches by user study.}
\vspace{-0.2em}
\centering
\resizebox{0.46\textwidth}{!}{%
\begin{tabular}{|c|c|c|c|}
\hline
\textbf{FGSM}~\cite{goodfellow2014explaining} & \textbf{PGD}~\cite{madry2017towards}  & \textbf{Proposed (image)} & \textbf{Proposed (text)} \\ \hline
 1.2$\%$    &  0$\%$   &  61$\%$   &   37.8$\%$  \\ \hline
\end{tabular}
}
\label{tab:user_noise}
\vspace{-0.5em}
\end{table}

\textbf{User Study:}  To verify that the adversarial faces generated using the proposed reference image-based approach are realistic and capture the desired attributes of the reference image, we conduct a user study. In the first experiment, we provided participants with 25 reference images and the corresponding three generated adversarial images (coarse, middle, and fine level) using our image-driven approach by optimizing latent codes and noise vector. Participants were asked to rate each of the three generated output images on a scale of 1 to 10 based on pose transfer (coarse level), expression transfer (middle level), and color transfer (fine level). We received 250 responses. The results indicate an average score of 9.1, which indicates the faithful transfer of the desired attributes at each level. We have not included other attributes, such as eyeglasses, hair color, etc., to make the study easy for the participants.
Similarly, we conduct a study to classify real vs. fake images. We provided participants with 50 images, half of which are real and half generated with StyleGAN2 without any perturbation. The participants labeled 81$\%$ of the StyleGAN generated images as real. The number is 72$\%$ for our reference image-based approach, showing only 9$\%$ drop in accuracy compared to unperturbed images generated by StyleGAN2. Similarly to compare with noise based methods, we provide 10 users with 50 quadruplet images consisting of the proposed image-based and text-based approaches, and two noise based methods (FGSM and PGD). The users are asked to select the most realistic face image. As shown in Table \ref{tab:user_noise}, about 61$\%$ of the users have selected the image generated via proposed approach.

\begin{figure}[t]
\centering
\includegraphics[trim={0cm 0cm 13.48cm 0},clip,width=0.48\textwidth]{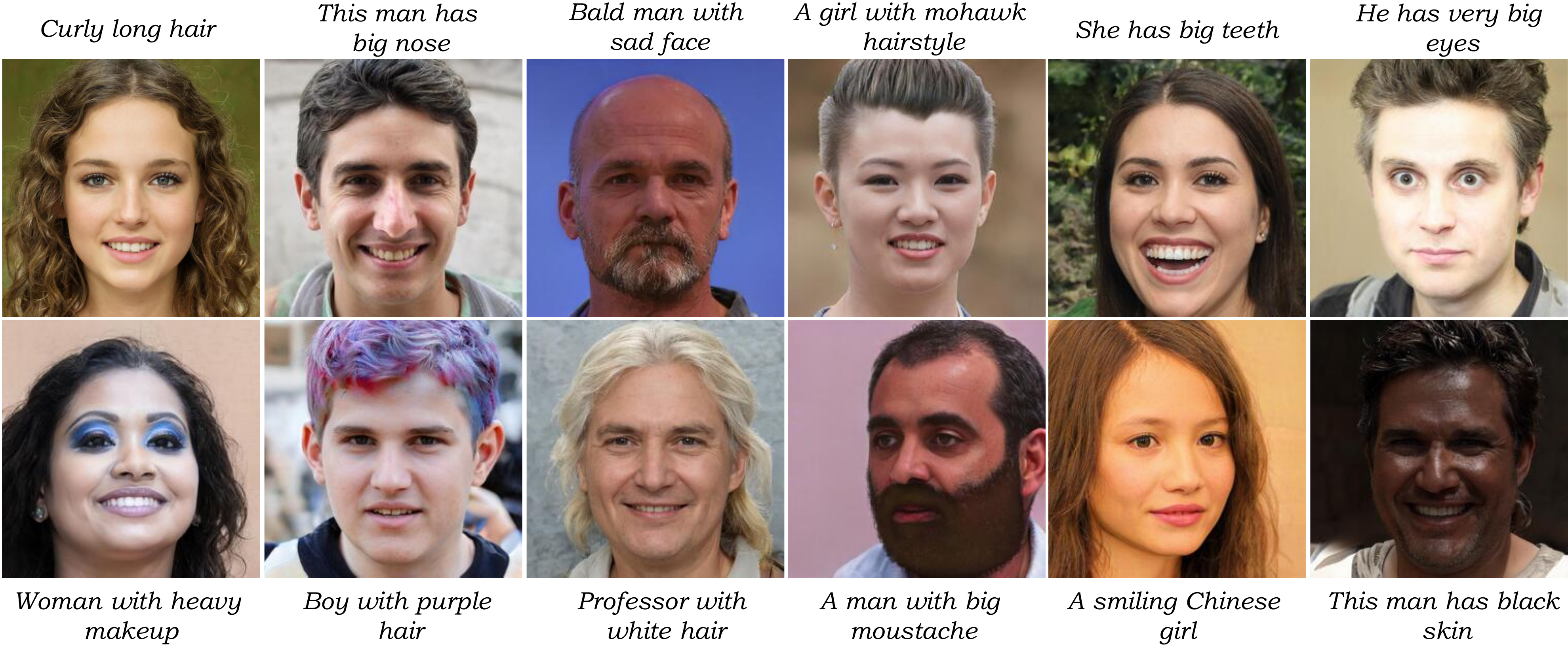}
\caption{Attribute-conditioned adversarial face images generated via our text-guided method. The driving text prompt is indicated next to the corresponding image. It can be seen that the generated images contain the attributes provided via the text prompt. Additionally, all these images are misclassified by forensic model (DenseNet-121).}
\label{fig:clip_qual}
\end{figure}

\subsection{Text as Reference:} \label{sec:textasref}
To evaluate the text-guided approach, we generate $50$ text prompts based on different facial features and ethnicities (details in the supplementary material).

We use Adam optimizer to jointly optimize noise and attribute-specific style vectors to minimize Eq. \ref{eq:2} with a learning rate of \koushik{0.001}. We set the total number of iterations to \koushik{50}.
Note that although the generated image from StyleGAN2 is always able to fool the target classifier, we have noticed that the match with the reference text depends on the initialization of image and weightage factors of CLIP loss and $\ell_2$ loss that controls the similarity with the initial image (see Eq. \ref{eq:2}). To provide a good initialization, we randomly sample 50 latent vectors and choose one that gives the lowest error between the generated image and the reference text in the CLIP embedding space.
The adversarial images generated using the text-guided approach are shown in Figure \ref{fig:clip_qual}. It can be seen that the proposed approach is capable of generating visually appealing adversarial images that match the provided text prompts. 

\begin{figure}[!t]
\centering
\includegraphics[width=0.11\textwidth]{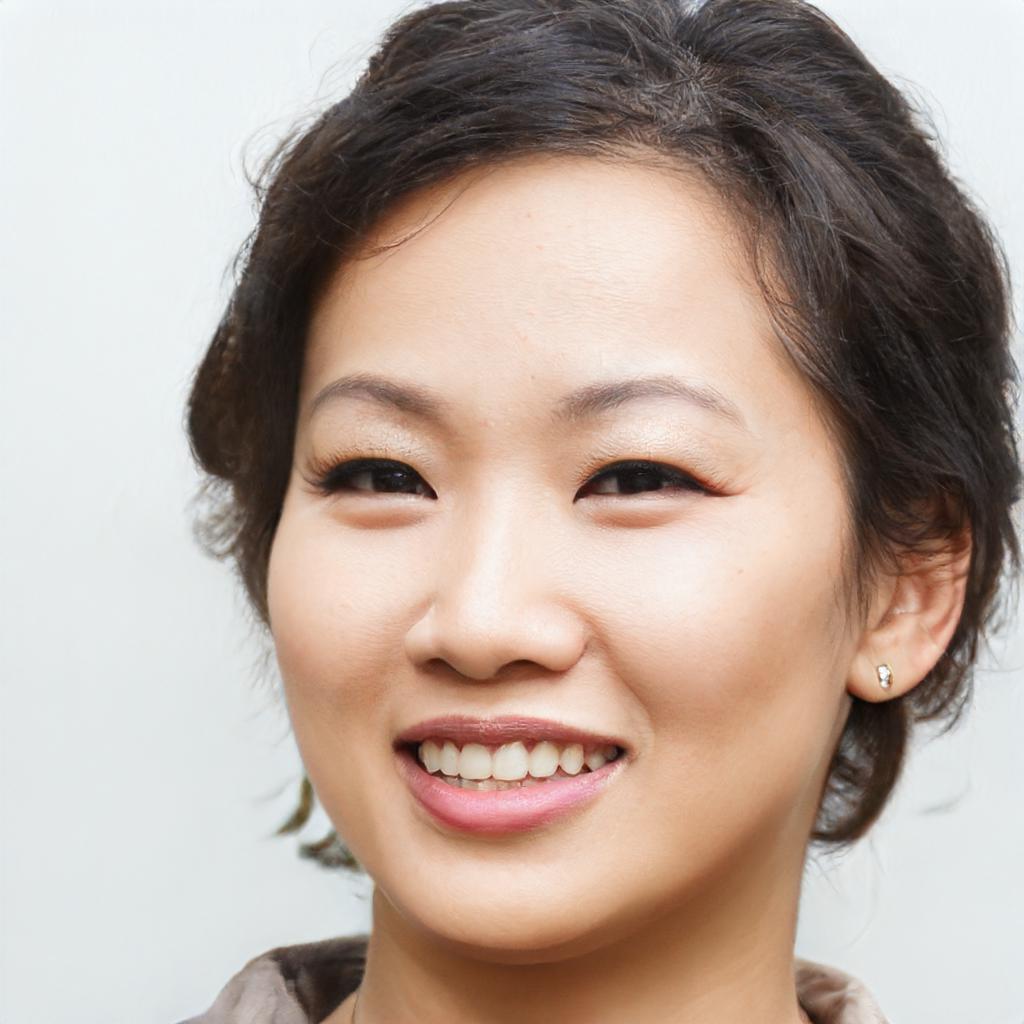}
\includegraphics[width=0.11\textwidth]{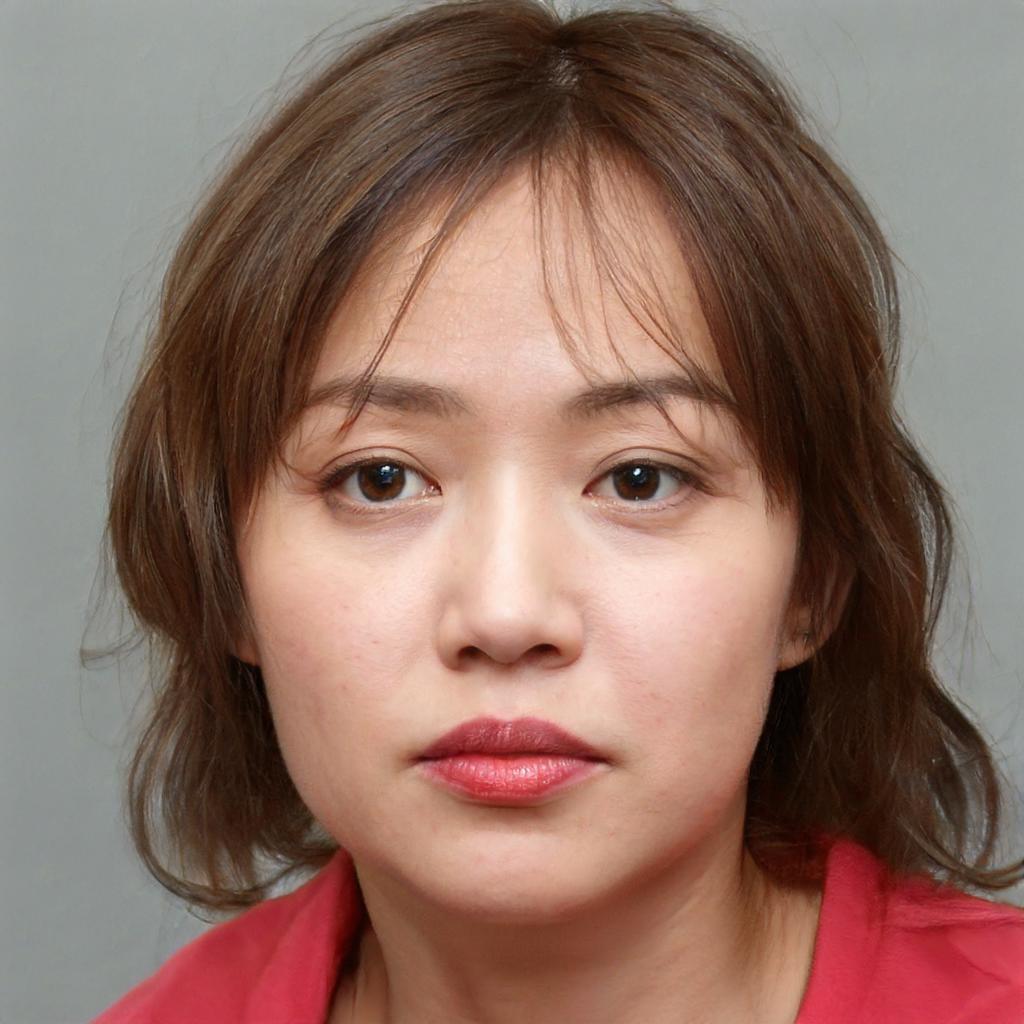}
\includegraphics[width=0.11\textwidth]{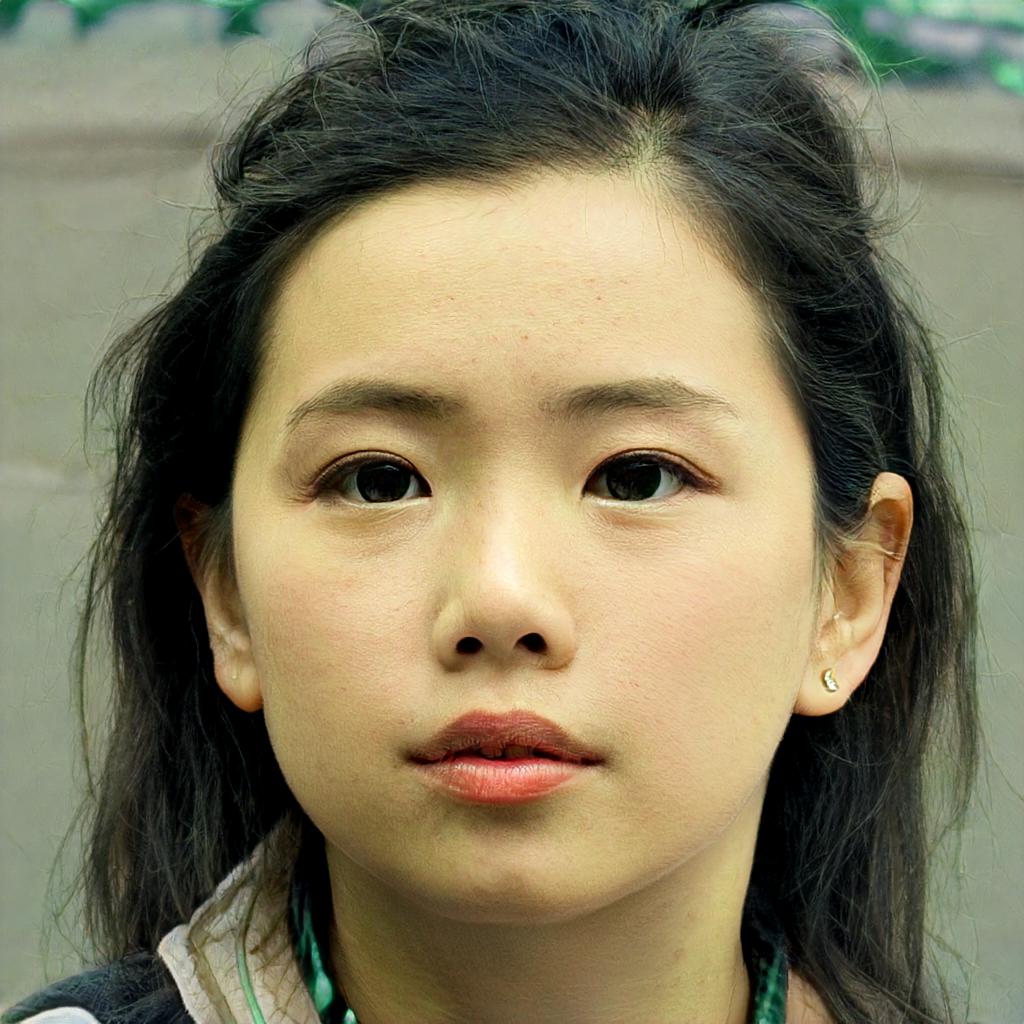}
\includegraphics[width=0.11\textwidth]{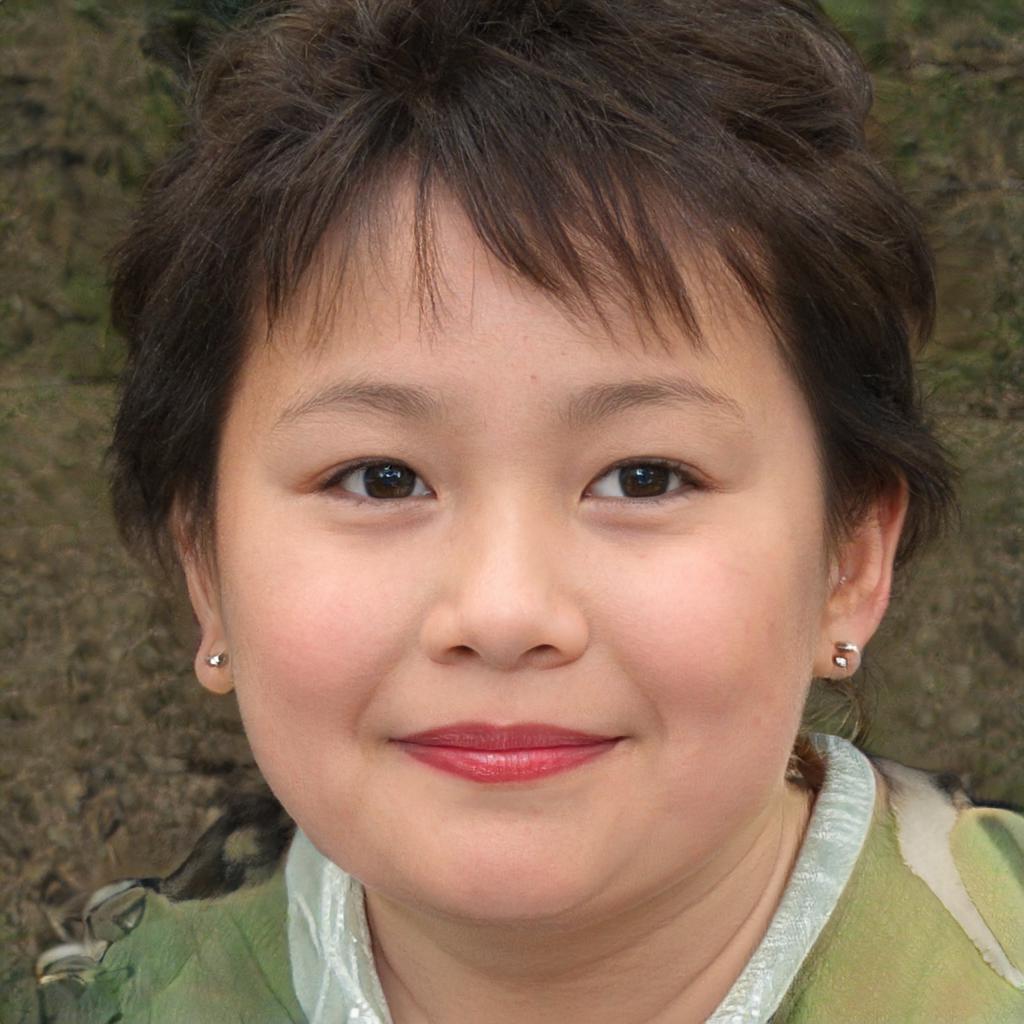}
\includegraphics[width=0.11\textwidth]{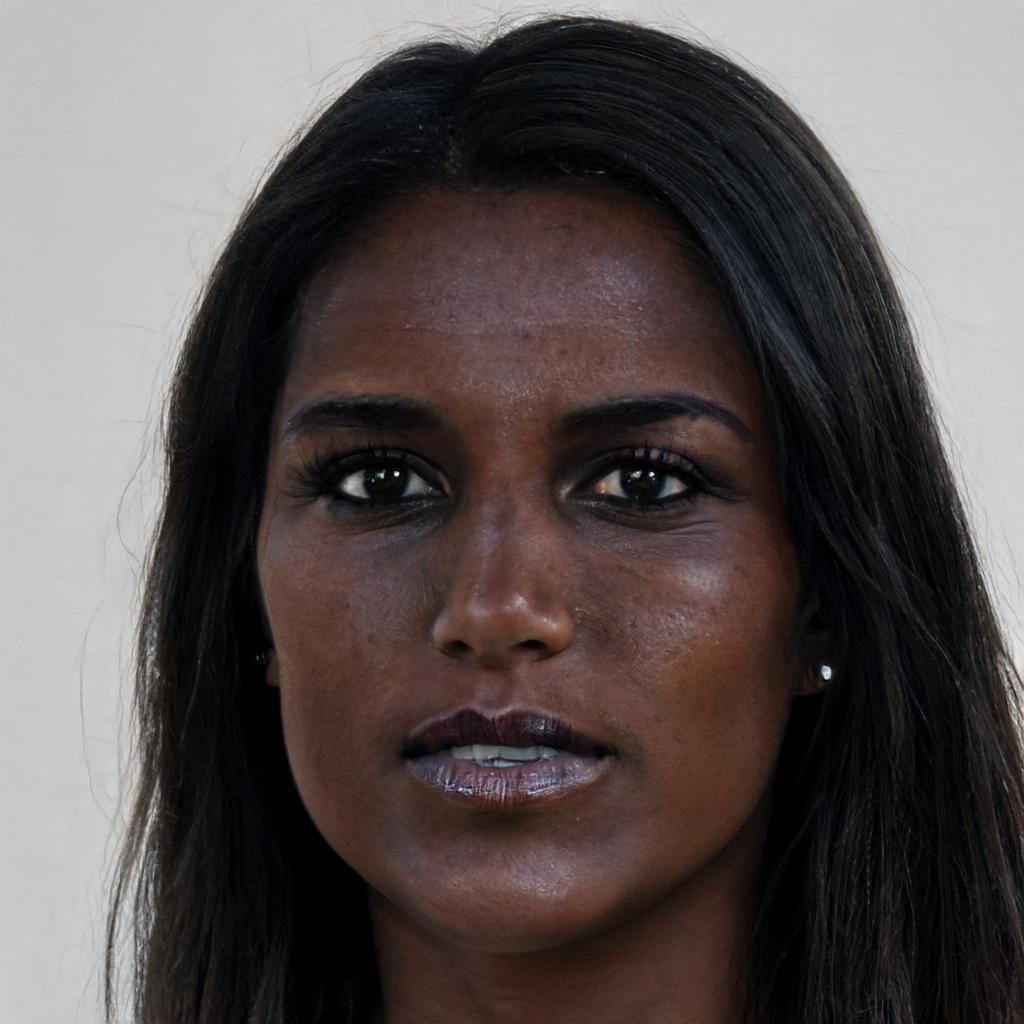}
\includegraphics[width=0.11\textwidth]{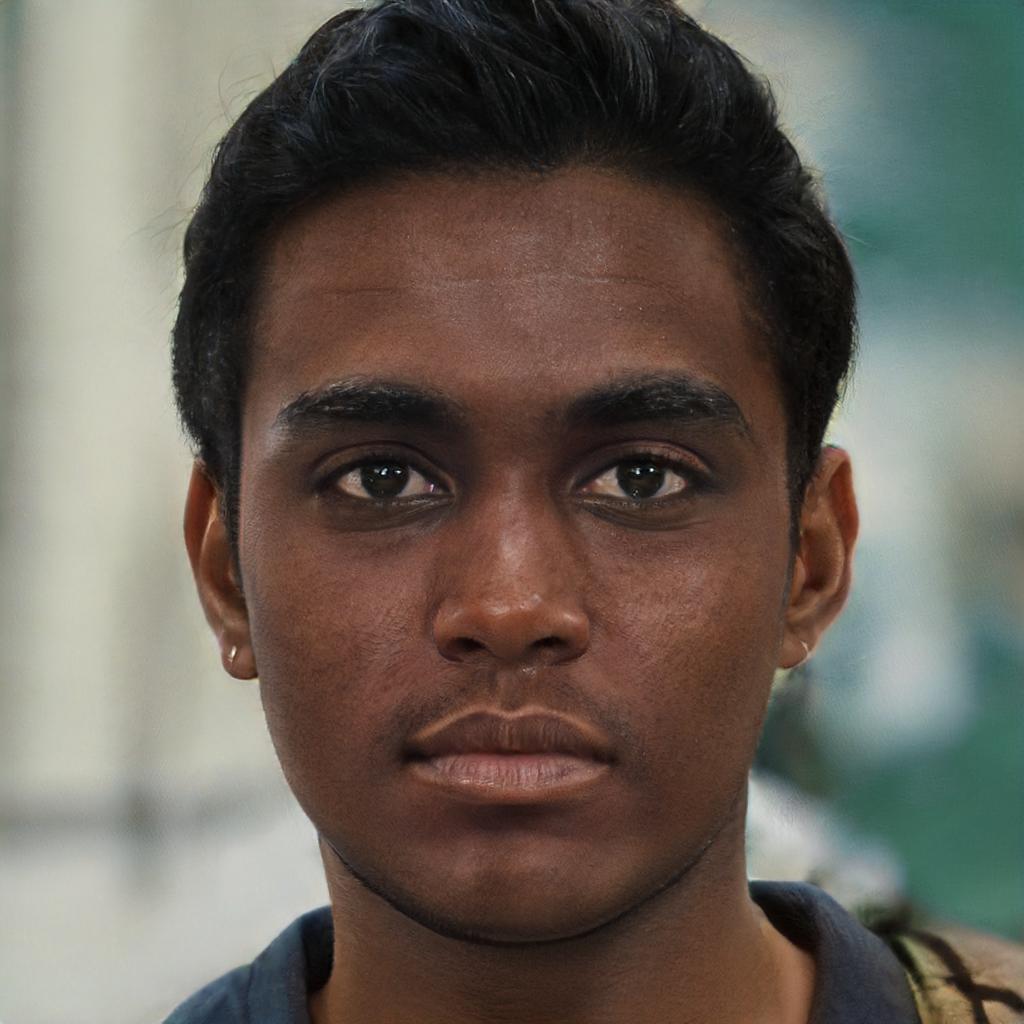}
\includegraphics[width=0.11\textwidth]{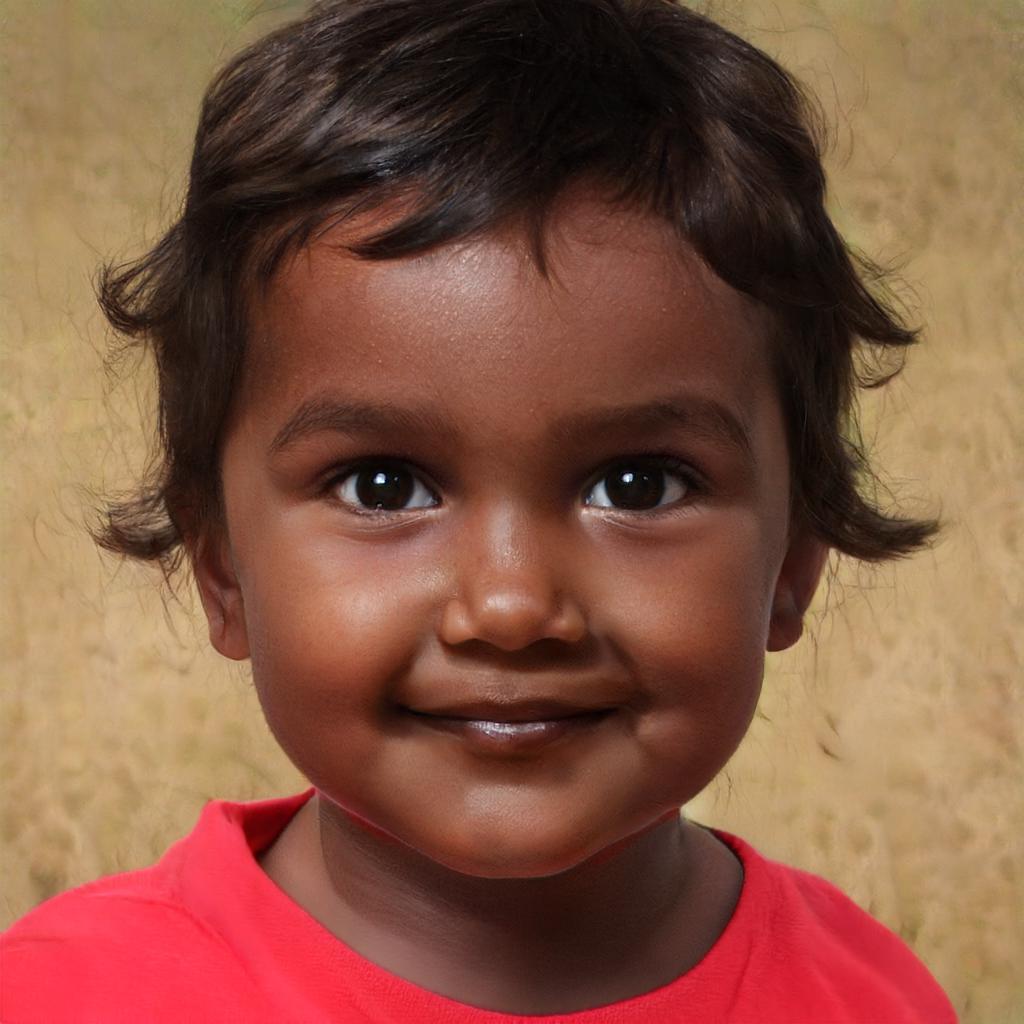}
\includegraphics[width=0.11\textwidth]{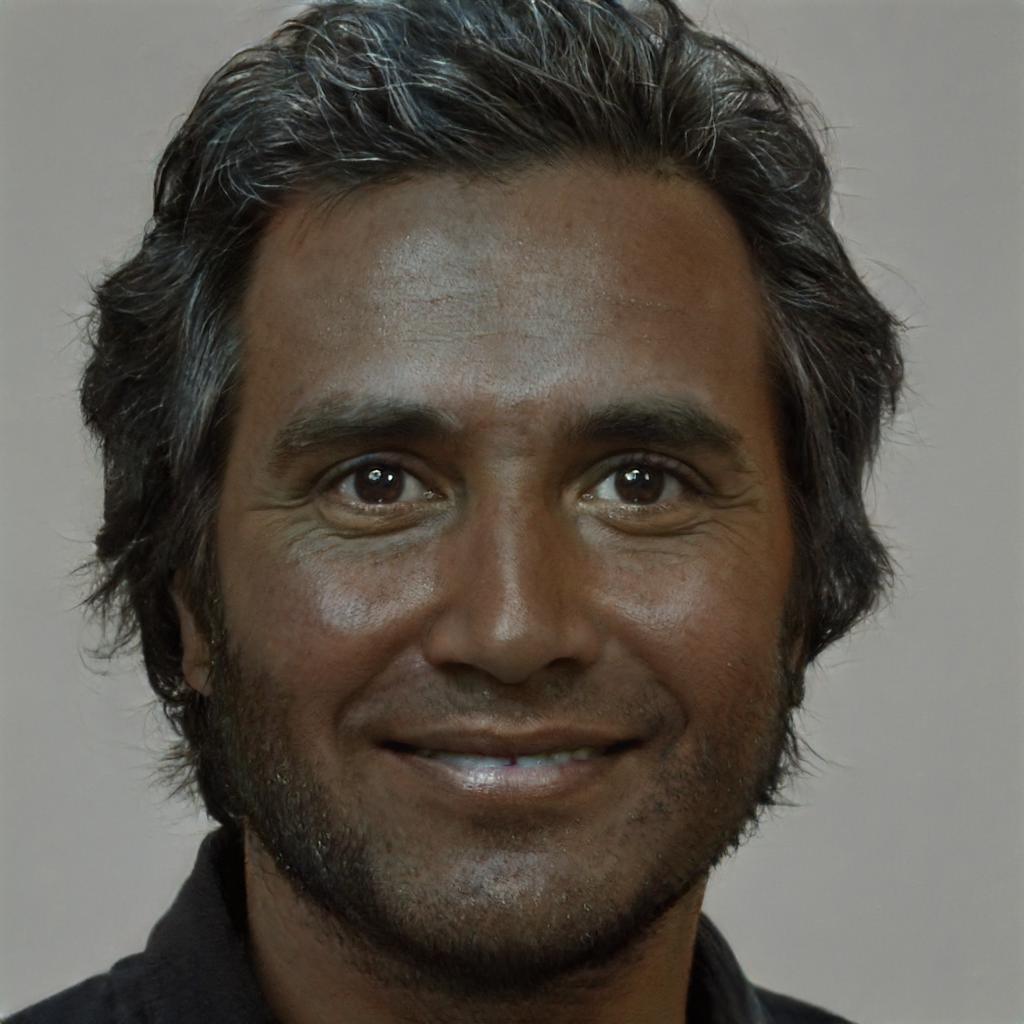}
\caption{Attackers can generate diverse adversarial images corresponding to a specific text prompt using the proposed text-guided approach to rapidly spread disinformation among specific ethnic groups. We show examples corresponding to text-prompt \texttt{Chinese girl} (top row) and \texttt{Dark skin} (bottom row). All generated images are misclassified by the forensic classifiers.}
\label{fig:diverse}
\vspace{-1.0em}
\end{figure}

Furthermore, the qualitative results in Table \ref{tab:classifiers} show that the generated images have a high success rate in fooling the forensic classifiers and a low FID score compared to norm-constrained noise-based attacks.
\begin{figure}[t]
\centering
\includegraphics[trim={0.6cm 0cm 0cm 0cm}, clip, width=0.46\textwidth]{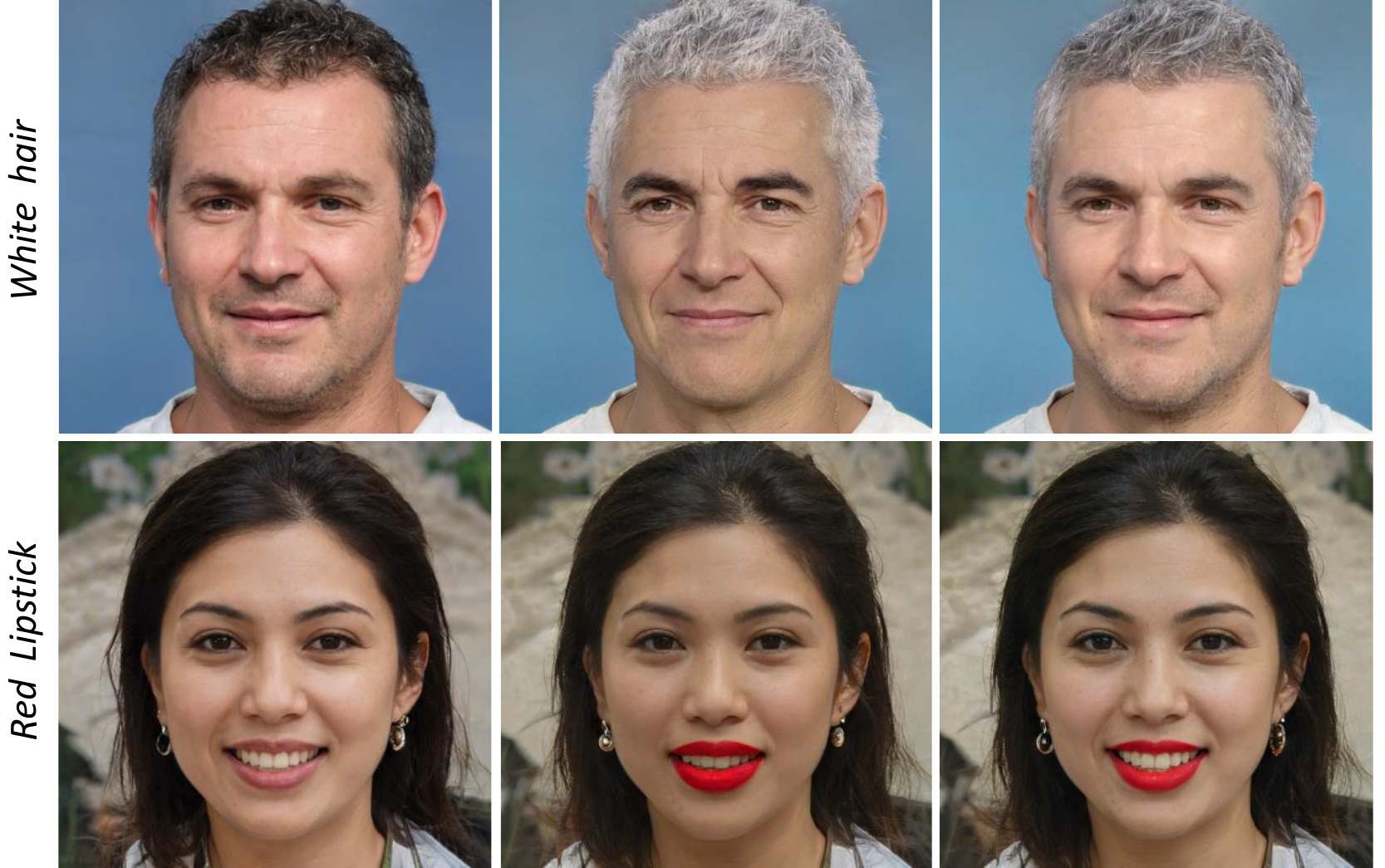}
\caption{\koushik{Effect of using the identity loss. First column shows the input image to be edited. Adversarial images generated with only Eq.\ref{eq:2} is shown in the second column. Adversarial image generated by adding identity loss to Eq.~\ref{eq:2} is shown in the third column. The driving text prompt for the top row is \textit{white hair} and for the bottom row is \textit{red lipstick}.}}
\label{fig:id_loss_change}
\end{figure}
\koushik{In Figure \ref{fig:id_loss_change}, we demonstrate the ability of our framework to preserve the identity of the input image without compromising its adversarial nature. This control is essential when we want to manipulate images in a real world setting. }
In Figure \ref{fig:diverse}, it can be seen that attackers can generate diverse adversarial images of any specific ethnic group using only a single text prompt using the proposed approach. Such a control can be used for malicious purposes and demands immediate attention from the image forensic researcher to develop an effective defense mechanism.
\textbf{User study:} As for the text-driven approach, we conducted a user study to evaluate the proximity of the synthesized samples generated by our text-guided approach to the provided text prompts. 
We select 15 generated adversarial face images along with the corresponding text prompts. The participants were asked to rate the similarity between the text prompt and the corresponding generated adversarial image via proposed approach on the scale of $1$ to $10$, where higher score indicate greater similarity. In total, we collect 250 responses with an average score of 8.5, indicating that the generated face images have attributes provided via text-prompt.
Similarly, we also conducted a human study to evaluate the quality of the generated face images. We collect 25 real images from the FFHQ dataset and 25 images generated by our text-guided approach and ask the users to label each image as real or fake. 69$\%$ of the generated images are labeled real by the users. 
\begin{table}
    \caption{\koushik{\textit{ASR} in the black box setting. All the models (except the one for which the score is reported) are used during optimization. We use the text-driven method as the base method in this case.}}
    \centering
    \footnotesize
    \setlength{\tabcolsep}{1.0pt}
    \renewcommand{\arraystretch}{1.3}
    \resizebox{0.46\textwidth}{!}
    %
    {
    \begin{tabular}{|c |c |c| c| c |c|}
    \hline
         \textbf{Method} & \textbf{ ResNet-18 } & \textbf{ ResNet-50 } & \textbf{ DenseNet-121 } & \textbf{ EfficientNet } & \textbf{ Xception } \\
        \hline
         \multirow{2}{*}{}{Ensemble}   & 11.0  & 32.0 & 54.0 & 46.0 & 11.0 \\
         \hline
         \multirow{2}{*}{}{Meta Learning}   & \textbf{12.0}  &\textbf{37.0}& \textbf{64.0} & \textbf{55.0} & \textbf{14.0} \\
        \hline
    \end{tabular}
    }

   \label{tab:blackbox_results}
    \vspace{-0.5em}
\end{table}
\subsection{Transferability of Adversarial Images}
\koushik{To evaluate the meta-learning based adversarial transferability described in section \ref{sec:metalearning}, we opt a leave-one-out strategy, where we leave out one forensic classifier to be used as the black box model and optimize the latents using the combination of losses from all the other classifiers. In Table \ref{tab:blackbox_results} we can see that combining the models in the meta-learning-based approach outperforms the ensemble-based method comfortably. This validates the idea of iterative simulating a black-box environment during optimization in the meta-test stage to increase the transferability to unknown models. Additional experiments under the meta-learning setting is provided in the supplementary material.}
\section{Conclusion}
In this paper, we have proposed a novel approach that leverages the latent space of the StyleGAN2 to generate image-driven and text-guided fake faces that are misclassified by forensic classifiers. We have conducted extensive experiments to show that faces generated using the proposed approach not only exhibit the desired attributes but are also able to fool the target forensic classifier with high success rate. We also show that our meta-learning-based approach is able to generate adversarial images that are transferable to unknown models.  We anticipate that building defenses against our proposed attribute based attack can shed light on the fragility of the current forensic classifiers, and eventually leads to more robust classifiers.\\

\noindent \underline{\textbf{Acknowledgement.}} The authors would like to thank Dr. Muzammal Naseer for several insightful discussions.

{\small
\bibliographystyle{ieee_fullname}
\bibliography{egbib}
}

\end{document}